\documentclass[10pt]{cai}

\usepackage[table]{xcolor}
\usetikzlibrary{positioning}
\definecolor{Gray}{gray}{0.9}
\definecolor{brockred}{RGB}{204,0,1}
\definecolor{softgreen}{RGB}{232,246,239}
\definecolor{softred}{RGB}{252,235,232}
\definecolor{softgold}{RGB}{255,247,220}
\definecolor{softblue}{RGB}{232,241,252}
\definecolor{softpurple}{RGB}{243,235,252}
\definecolor{emailicon}{RGB}{26,115,232}
\definecolor{githubicon}{RGB}{110,64,193}

\newcommand{\ours}{RAUL}
\newcommand{\oursuniform}{UniDist-RAUL}
\newcommand{\oursref}{HeldDist-RAUL}

\newcommand{\retrain}{{\text{Retrain}}}
\newcommand{\FT}{{\text{FT}}}
\newcommand{\GA}{{\text{GA}}}
\newcommand{\IU}{{\text{IU}}}
\newcommand{\RL}{{\text{RL}}}
\newcommand{\MUSparse}{{\text{$\ell_1$-sparse}}}

\begin{document}

\def\conferenceyear{2026}
\volumeheader{37}{0}
\begin{center}

\title{Multi-Objective Reference-Aligned Machine Unlearning}
\maketitle

\thispagestyle{empty}

\begin{tabular}{cc}
Rasa Khosrowshahli\upstairs{\affilone,*}, Stephen Asobiela\upstairs{\affilone}, Beatrice Ombuki-Berman\upstairs{\affilone}, Shahryar Rahnamayan\upstairs{\affiltwo}
\\[0.25ex]
{\small \upstairs{\affilone} Department of Computer Science, Brock University, St. Catharines, ON, Canada} \\
{\small \upstairs{\affiltwo} Yousef Haj-Ahmad Department of Engineering, Brock University, St. Catharines, ON, Canada} \\
\end{tabular}
  
\emails{
  \upstairs{*}rkhosrowshahli@brocku.ca (corresponding)
}
\vspace*{0.2in}
\end{center}

\begin{abstract}
Machine unlearning aims to remove the influence of specific training samples while preserving the model's utility. Existing single-objective approaches, such as gradient ascent or random relabeling, often induce catastrophic forgetting due to conflicting optimization dynamics and unbounded forgetting objectives that cause the model to drift from its pre-trained knowledge. We propose Reference-Aligned UnLearning (RAUL), a multi-objective framework that jointly optimizes forgetting and retention by replacing unbounded loss maximization with a bounded KL alignment of predictions on forgotten samples toward a reference distribution representing unseen data, instantiated either as a uniform distribution or an empirical distribution from a held-out reference set, which constrains the forgetting objective and reduces gradient conflict with retention. The resulting multi-objective optimization (MOO) problem is solved via Jacobian descent, which aggregates multiple gradients into a direction that does not conflict. Our results demonstrate that RAUL achieves the closest gap compared to full retraining.
\end{abstract}

\copyrightnotice

\section{Introduction}
Machine unlearning (MU) removes the influence of designated training points from a deployed model, motivated by privacy regulations~\cite{gdpr2016} and the prohibitive cost of retraining from scratch. Approximate methods usually optimize a single forgetting signal, which often conflicts with retention and yields catastrophic forgetting~\cite{CatastrophicForgetting}. The bottleneck is explicit conflict between forgetting and retaining~\cite{zhao2024makes}.

We address this unlearning challenge by proposing a multi-objective RAUL \footnote{Code available at \url{https://github.com/rkhosrowshahi/MO-RAUL}.} formulation over forgetting and retention~\cite{MOO-LLMUnlearning} that replaces loss maximization or random labeling on forgotten samples with a reference-based objective in which predictions on the forget set align with a reference distribution to simulate ``unseen'' data behavior. We propose two references, namely uniform and held-out data distribution. Together, they yield controllable forgetting by aligning the model's confidence on forget data with that of unseen data~\cite{cheng2024remainingdatafreemachineunlearningsuppressing}.

\section{Background and Related Works}
We first formalize the machine unlearning problem, then review gradient-based methods, and finally survey recent multi-objective (MO) formulations that motivate our approach.

\subsection{Problem Formulation: Machine Unlearning}
\label{sec:problem}

Let $\mathcal{D} = \mathcal{R} \cup \mathcal{F}$ denote the training dataset, where $\mathcal{F}$ represents the forget set containing samples requested for removal, and $\mathcal{R}$ denotes the retain set consisting of all remaining data. Given a pretrained model $f_{\theta}$ with parameters $\theta$ trained on $\mathcal{D}$, the goal of MU is to obtain an updated model $f_{\theta_u}$ such that the influence of samples in $\mathcal{F}$ is effectively removed from the model, and performance on $\mathcal{R}$ is preserved or minimally degraded.

\subsection{Gradient-Based Unlearning}
Gradient ascent (\GA) is widely used as a baseline, and it maximizes loss on the forget set so that the model is pushed toward misclassifying those inputs~\cite{UnlearnPretrainedLLM}. A related strategy is to fine-tune while replacing supervision on $\mathcal{F}$ with random labels~\cite{Li2023RandomRelabeling}. Together, these approaches cast forgotten samples as adversarial targets to be repelled from correct predictions, which frequently conflicts with the retention objective and can produce catastrophic forgetting and sharp utility loss on $\mathcal{R}$~\cite{CatastrophicForgetting,GSLoRApracticalforgetting}. Subsequent work seeks to ease this tension by projecting forgetting gradients so they conflict less with retention gradients. Such refinements improve stability but still treat unlearning on $\mathcal{F}$ as untraining.

Widely used benchmarks mix several instantiations of this tension. {\GA}~\cite{thudi2022unrolling} and {\RL}~\cite{golatkar2020eternal} foreground forgetting by increasing loss or randomizing labels on $\mathcal{F}$. {\FT} instead omits $\mathcal{F}$ from continued training on $\mathcal{R}$~\cite{warnecke2021machine}, which stabilizes updates but may leave residual influence from the pretrained snapshot. {\IU} and {\MUSparse} pursue approximate removal through influence-based and sparsity-constrained parameter updates~\cite{izzo2021approximate,jia2023model}, while SalUn~\cite{SalUn} couples saliency with forget-set adaptation under a shared training recipe. Despite their diversity, these pipelines often still tie forgetting to aggressive error shaping on forgotten data or to localized edits that interfere with retained representations. 

\subsection{Multi-Objective Unlearning}
Recent work treats forgetting and retention as competing objectives; Wu and Harandi~\cite{MUNBAUnlearning} formulate MU as Nash bargaining with Pareto guarantees, but careful utility design and computational cost limit scalability to large pretrained models.

\section{Reference-Aligned UnLearning}
\begin{figure}[t]
\centering
\resizebox{\linewidth}{!}{%
\begin{tikzpicture}[
  transform shape,
  node distance=0.62cm and 0.58cm,
  model/.style={draw=brockred!45, fill=softblue, circle, align=center, minimum size=1.2cm, font=\scriptsize},
  objbox/.style={rounded corners, align=center, minimum height=0.9cm, text width=3.45cm, minimum width=3.45cm, font=\scriptsize},
  retain/.style={objbox, draw=green!45!black, fill=softgreen},
  forget/.style={objbox, draw=brockred!70, fill=softred},
  op/.style={draw=brockred!65, fill=softgold, rounded corners, align=center, minimum height=1.15cm, minimum width=2.20cm, font=\scriptsize},
  update/.style={draw=brockred!65, fill=softgold, rounded corners, align=center, minimum height=1.15cm, minimum width=2.15cm, font=\scriptsize},
  output/.style={draw=brockred!45, fill=softblue, circle, align=center, minimum size=1.2cm, font=\scriptsize},
  note/.style={font=\scriptsize\bfseries, color=brockred, align=center},
  arrow/.style={->, very thick, color=brockred},
  objarrow/.style={->, very thick},
  looparrow/.style={->, thick, dashed, color=brockred!75}
]
  \node[model] (model) {Pretrained\\$f_\theta$};
  \node[retain, right=of model, xshift=-0.4cm, yshift=0.55cm] (retain) {\textbf{Retain} on $\mathcal{R}$\\[0.12em]fit $p_\theta(y_{\mathcal{R}}\mid x_{\mathcal{R}})$ to $y_{\mathcal{R}}$\\[0.05em] $g_{\mathcal{R}}=\nabla_\theta\mathcal{L}_{\text{retain}}$};
  \node[forget, right=of model, xshift=-0.4cm, yshift=-0.55cm] (forget) {\textbf{Forget} on $\mathcal{F}$\\[0.12em]align $p_\theta(y\mid x_{\mathcal{F}})$ to $p_{\text{ref}}(y)$\\[0.05em] $g_{\mathcal{F}}=\nabla_\theta\mathcal{L}_{\text{forget}}$};
  \node[op, right=of retain, yshift=-0.55cm] (upgrad) {\textbf{UPGrad} (MOO)\\[0.5em]$J=\bigl[g_{\mathcal{R}},\,g_{\mathcal{F}}\bigr]^{\top}$\\$d=\mathcal{A}(J)$};
  \node[update, right=of upgrad] (step) {\textbf{Update}\\[0.5em]$\theta^{t+1}=\theta^t-\eta d$};
  \node[output, right=of step] (out) {Unlearned\\$f_{\theta_u}$};
  \node[note, above=0.18cm of upgrad, font=\small] {Multi-objectives $\rightarrow$ \\ Pareto descent};
  \draw[objarrow, color=green!45!black] (model.east) -- (retain.west);
  \draw[objarrow, color=brockred] (model.east) -- (forget.west);
  \draw[objarrow, color=green!45!black] (retain.east) -- (upgrad.west);
  \draw[objarrow, color=brockred] (forget.east) -- (upgrad.west);
  \draw[arrow] (upgrad) -- (step);
  \draw[arrow] (step) -- (out);
  \draw[looparrow] (step.south) .. controls +(0,-1.5cm) and +(0,-1.5cm) ..
    node[below, yshift=-0.08cm, font=\small, align=center] {Iteration} (model.south);
\end{tikzpicture}%
}
\vspace{0.2em}
\scriptsize
\noindent
\setlength{\fboxsep}{1pt}
\setlength{\fboxrule}{0.7pt}
\begin{minipage}[t]{0.39\linewidth}
\colorbox{softblue}{%
\parbox[t]{0.99\linewidth}{%
\centering
\textbf{Uniform Dist. (UniDist)}\par\vspace{-0.65em}
{\color{emailicon!80!black}\rule{0.92\linewidth}{0.55pt}}\par
\raggedright
\textbf{Set:} $p_{\text{ref}}(y)=1/C$.\\
\textbf{Effect:} predictions on $\mathcal{F}$ become uncertain.%
}%
}
\end{minipage}\hspace{0.5em}%
\begin{minipage}[t]{0.54\linewidth}
\colorbox{softpurple}{%
\parbox[t]{0.99\linewidth}{%
\centering
\textbf{Heldout Dist. (HeldDist)}\par\vspace{-0.65em}
{\color{githubicon!85!black}\rule{0.92\linewidth}{0.55pt}}\par\vspace{-0.2em}
\raggedright
\textbf{Set:} $p_{\text{ref}}(y)$ from disjoint held-out $\mathcal{D}_{\text{ref}}$.\\
\textbf{Effect:} predictions on $\mathcal{F}$ mimic unseen data behavior with better retention stability.%
}%
}
\end{minipage}
\caption{Overview of the proposed {\ours} framework. The retain objective preserves performance on $\mathcal{R}$, while the forgetting objective aligns predictions on $\mathcal{F}$ with a reference distribution. UPGrad aggregates the two objective gradients into a Pareto descent update.}
\label{fig:raul-overview}
\end{figure}
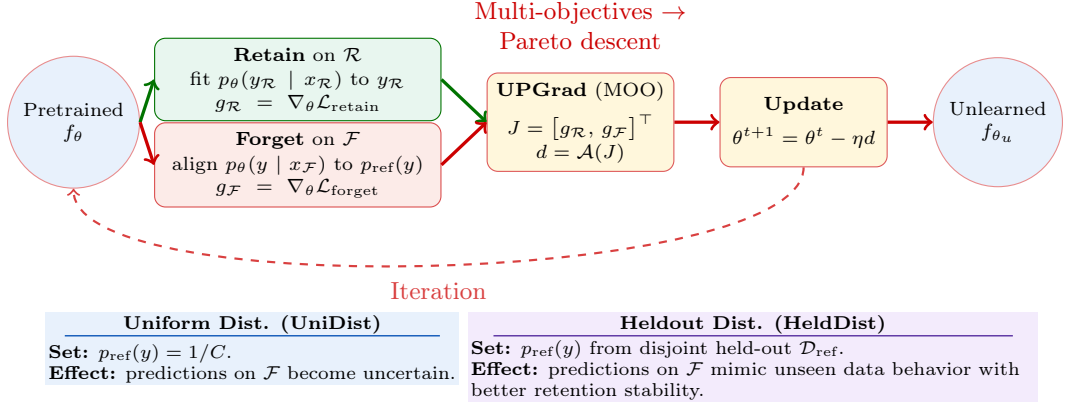

We propose multi-objective \textbf{R}eference-\textbf{A}ligned \textbf{U}n\textbf{L}earning ({\ours}), a framework that formulates machine unlearning as a two objective optimization problem. Traditional methods that maximize error on forgotten samples push the model to strongly reject those inputs, damaging its internal structure and causing it to forget things it should have kept~\cite{CatastrophicForgetting}. Rather than adversarially repelling forgotten samples, we align their predictions with a reference distribution representing unseen data, while simultaneously minimizing retention loss on the remaining data, so that a truly forgotten sample yields predictions similar to what the model would produce for data it has never encountered. We formally define the two objectives, detail the two reference distribution strategies, and present the MOO formulation solved via Jacobian descent.
Figure~\ref{fig:raul-overview} summarizes the two-objective unlearning workflow and the two reference distributions used by {\ours}.

\subsection{Retention Objective}
For the retention objective, we minimize the standard cross-entropy loss on $\mathcal{R}$ as:
\begin{equation}
\min_{\theta} \; \mathcal{L}_{\text{retain}}(f_{\theta}, \mathcal{R}) = -\mathbb{E}_{(x, y) \sim \mathcal{R}} \left[ \log p_{\theta}(y \mid x) \right],
\label{eq:retention_objective}
\end{equation}
where $p_{\theta}(y \mid x)$ is the model's predicted probability for the true label $y$ given input $x$. This objective maintains predictive performance on $\mathcal{R}$ during unlearning.

\subsection{Reference-based Forgetting Objective}
We propose a Reference-based forgetting objective that aligns the model's predictive distribution on forgotten samples with a suitable reference distribution. Formally, we minimize the KL divergence
\begin{equation}
\min_{\theta} \; \mathcal{L}_{\text{forget}}(f_{\theta}, \mathcal{F}) =
\mathbb{E}_{x_u \sim \mathcal{F}} \left[ \mathrm{KL}\left(
p_{\theta}(y \mid x_u)
\;\|\;
p_{\text{ref}}(y)
\right) \right],
\label{eq:reference_forgetting}
\end{equation}
where $p_{\theta}(y \mid x_u)$ denotes the model's predictive distribution on forget sample $x_u$, and $p_{\text{ref}}(y)$ is a reference distribution representing unseen data.

The choice of reference distribution is crucial. We propose two strategies. We either use a uniform distribution as a principled baseline or leverage an unseen reference dataset, $\mathcal{D}_{\text{ref}}$, to improve stability. When a target dataset is used as reference, we compute the reference distribution as the expected predictive distribution over the samples as $p_{\text{ref}}(y) = \mathbb{E}_{x' \sim \mathcal{D}_{\text{ref}}} \left[ p_{\theta}(y \mid x') \right]$.
This approach aligns the model's predictions on forgotten samples to exhibit uncertainty similar to truly unseen data.

\subsection{Reference Distributions}
\label{subsec:ref-distributions}
\textbf{Uniform Distribution (UniDist).} \ \ The uniform distribution serves as a principled baseline requiring no additional data. Setting $p_{\text{ref}}(y) = 1/C$, where $C$ is the number of outputs, encourages the model to treat forgotten samples with maximum uncertainty, as if it has no prior knowledge about them. Notably, minimizing the KL divergence to a uniform distribution is equivalent to maximizing the entropy of the predicted distribution, since $\mathrm{KL}(p_{\theta}(y \mid x_u) \| U) = \log C - H(p_{\theta}(y \mid x_u))$, where $H(\cdot)$ denotes entropy. This objective is bounded between 0 and $\log C$, providing stable optimization dynamics.

\textbf{Held-out Distribution (HeldDist).} \ \ We compute an empirical reference distribution from an unseen held-out set $\mathcal{D}_{\text{ref}}$ that was not part of the original training. This set should be disjoint from $\mathcal{R}$, $\mathcal{F}$, and any test or validation sets to represent a truly unseen distribution and avoid information leakage during evaluation. Using such a reference provides a more stable anchor that better preserves learned representations on $\mathcal{R}$.

\subsection{Multi-Objective Alignment in Unlearning}
We recall the multi-objective notions on which our analysis relies. For a vector objective $F(\theta) = (f_1(\theta), \ldots, f_m(\theta))$ over parameter space $\Theta$, we say that $\theta_a$ dominates $\theta_b$ when $f_i(\theta_a) \le f_i(\theta_b)$ for all $i$ and $f_j(\theta_a) < f_j(\theta_b)$ for at least one $j$. A point $\theta^\star$ is Pareto-optimal when no other point in $\Theta$ dominates it, and the image of all Pareto-optimal points under $F$ is the Pareto front. For smooth non-convex problems, the relevant first-order condition is Pareto stationarity. A point $\theta$ is Pareto-stationary when there exist non-negative weights $\alpha_1, \ldots, \alpha_m$ with $\sum_i \alpha_i = 1$ that satisfy $\sum_{i=1}^{m} \alpha_i \nabla_\theta f_i(\theta) = 0$. Pareto stationarity is a necessary condition for Pareto optimality and is the strongest guarantee typically available for gradient-based MO methods such as Multiple Gradient Descent Algorithm (MGDA)~\cite{desideri2012mgda,sener2018multitask} and Jacobian descent~\cite{JacobianDescent} on deep networks.

We transform {\ours} into the bi-objective problem
\begin{equation}
\theta_u = \arg\min_{\theta} \left( \mathcal{L}_{\text{retain}}(f_{\theta}, \mathcal{R}), \; \mathcal{L}_{\text{forget}}(f_{\theta}, \mathcal{F}) \right),
\label{eq:joint_objective}
\end{equation}
in which both objectives are optimized jointly rather than collapsed into a single scalar. In this setting $m=2$, so a Pareto-stationary parameter satisfies $\alpha_r g_r + \alpha_f g_f = 0$ for some $\alpha_r, \alpha_f \ge 0$ with $\alpha_r + \alpha_f = 1$, where $g_r = \nabla_\theta \mathcal{L}_{\text{retain}}$ and $g_f = \nabla_\theta \mathcal{L}_{\text{forget}}$. The two objectives are designed to be in conflict, so the inner product $\langle g_r, g_f \rangle$ is negative throughout training, and any unprojected combination of the two gradients can ascend one objective while descending the other. To solve Eq.~\ref{eq:joint_objective} we employ Jacobian descent~\cite{JacobianDescent}, which operates directly on the per-objective gradients without requiring a manual trade-off parameter. 
We stack the per-objective gradients into a Jacobian matrix and compute an aggregated update direction as
\begin{equation}
J = \begin{bmatrix} g_r^T \\ g_f^T \end{bmatrix}, \qquad G = \mathcal{A}(J),
\label{eq:jacobian_aggregate}
\end{equation}
where $\mathcal{A}$ is an aggregator function that aggregates gradients into a single gradient $G$ to update the parameters at step $t$ with learning rate $\eta$ as
\begin{equation}
\theta^{t+1}_u = \theta^{t}_u - \eta\, G^{t},
\label{eq:param_update}
\end{equation}

By contrast, the classic linear scalarization (LS) with a fixed weight $\lambda$ drives the parameters toward a stationary point of the weighted sum $\lambda \mathcal{L}_{\text{retain}} + (1-\lambda) \mathcal{L}_{\text{forget}}$. The resulting update depends on the fixed weight $\lambda$; it may still carry a conflicting component when $g_r$ and $g_f$ oppose each other, and generally cannot reach the non-convex portion of the Pareto front.
We use the Unconflicting Projection of Gradients (UPGrad) aggregator $\mathcal{A}_{\text{UPGrad}}$~\cite{JacobianDescent}, which projects each objective gradient onto the dual cone of all objective gradients and then averages the projected vectors. 
The fixed points of $\mathcal{A}_{\text{UPGrad}}$ therefore coincide with Pareto-stationary points of Eq.~\ref{eq:joint_objective}. 
Utilizing multiple points on the Pareto front is outside the scope of this work and motivates the use of multi-objective evolutionary algorithms \cite{deb2002fast} for unlearning as future work.

\section{Experiments}
We evaluate the effectiveness of our proposed {\ours} framework through extensive experiments on image classification, comparing against representative unlearning baselines under varying forgetting ratios. Full details of the experimental setup are provided in Appendix~\ref{app:setup}.

\begin{table}[h]
    \centering
    \caption{Comparison of unlearning methods on CIFAR-10 with ResNet-18 under 10\% and 50\% random data forgetting. Results show $mean_{\pm std}$ over all runs in each forgetting setting. Values in parentheses indicate the performance gap to {\retrain} (lower is better).}
    \label{tab: classification_data_ratio}
    \vspace*{-2mm}
    \resizebox{\textwidth}{!}{%
    \setlength{\tabcolsep}{4pt}%
    \begin{tabular}{c|ccccc}
    \toprule[1pt]
    \midrule
    \multicolumn{6}{c}{\textbf{Random 10\%}} \\
    \midrule
    \textbf{Methods} & \multicolumn{1}{c|}{UA $\uparrow$} & \multicolumn{1}{c|}{RA $\uparrow$} & \multicolumn{1}{c|}{TA $\uparrow$} & \multicolumn{1}{c|}{MIA $\uparrow$} & \multicolumn{1}{c}{Avg. Gap $\downarrow$} \\
    \midrule
    \rowcolor{white}
    {{\retrain}} & ${5.68}_{\pm0.27}$ (\textcolor{blue}{0.00}) & ${100.00}_{\pm0.00}$ (\textcolor{blue}{0.00}) & ${94.32}_{\pm0.03}$ (\textcolor{blue}{0.00}) & ${13.13}_{\pm1.16}$ (\textcolor{blue}{0.00}) & \textcolor{blue}{0.00} \\
    \midrule
    \FT & ${3.93}_{\pm5.53}$ (\textcolor{blue}{5.06}) & ${97.56}_{\pm3.48}$ (\textcolor{blue}{2.44}) & ${92.22}_{\pm3.32}$ (\textcolor{blue}{2.47}) & ${8.15}_{\pm9.98}$ (\textcolor{blue}{23.79}) & \textcolor{blue}{8.44} \\
    \GA & ${11.29}_{\pm31.73}$ (\textcolor{blue}{15.56}) & ${88.73}_{\pm31.75}$ (\textcolor{blue}{11.27}) & ${84.02}_{\pm29.86}$ (\textcolor{blue}{10.84}) & ${1.93}_{\pm3.33}$ (\textcolor{blue}{29.93}) & \textcolor{blue}{16.90} \\
    \IU & ${0.74}_{\pm1.17}$ (\textcolor{blue}{5.13}) & ${99.15}_{\pm1.28}$ (\textcolor{blue}{0.85}) & ${93.32}_{\pm1.69}$ (\textcolor{blue}{1.26}) & ${1.95}_{\pm2.12}$ (\textcolor{blue}{17.00}) & \textcolor{blue}{6.06} \\
    \MUSparse & ${56.12}_{\pm42.94}$ (\textcolor{blue}{53.00}) & ${44.04}_{\pm42.94}$ (\textcolor{blue}{55.96}) & ${41.98}_{\pm40.09}$ (\textcolor{blue}{52.42}) & ${21.43}_{\pm21.25}$ (\textcolor{blue}{15.22}) & \textcolor{blue}{44.15} \\
    \RL & ${18.75}_{\pm24.47}$ (\textcolor{blue}{18.32}) & ${82.38}_{\pm24.85}$ (\textcolor{blue}{17.62}) & ${78.75}_{\pm22.22}$ (\textcolor{blue}{15.69}) & ${32.28}_{\pm30.04}$ (\textcolor{blue}{21.58}) & \textcolor{blue}{18.30} \\
    SalUn & ${7.68}_{\pm15.91}$ (\textcolor{blue}{10.28}) & ${93.07}_{\pm15.45}$ (\textcolor{blue}{6.93}) & ${88.23}_{\pm13.59}$ (\textcolor{blue}{6.21}) & ${14.75}_{\pm15.89}$ (\textcolor{blue}{11.48}) & \textcolor{blue}{8.72} \\
    \midrule
    \rowcolor{Gray}
    \textbf{\oursuniform-LS} & ${0.97}_{\pm0.99}$ (\textcolor{blue}{4.52}) & ${99.20}_{\pm1.02}$ (\textcolor{blue}{0.80}) & ${91.56}_{\pm2.97}$ (\textcolor{blue}{2.78}) & ${7.44}_{\pm0.49}$ (\textcolor{blue}{4.87}) & \textcolor{blue}{3.24} \\
    \rowcolor{Gray}
    \textbf{\oursuniform-UPGrad} & ${1.43}_{\pm1.41}$ (\textcolor{blue}{4.06}) & ${99.11}_{\pm1.38}$ (\textcolor{blue}{0.89}) & ${91.95}_{\pm2.77}$ (\textcolor{blue}{2.39}) & ${9.52}_{\pm0.72}$ (\textcolor{blue}{2.79}) & \textcolor{blue}{2.53} \\
    \rowcolor{Gray}
    \textbf{\oursref-LS} & ${7.30}_{\pm2.71}$ (\textcolor{blue}{2.97}) & ${99.24}_{\pm0.16}$ (\textcolor{blue}{0.76}) & ${92.09}_{\pm0.48}$ (\textcolor{blue}{2.25}) & ${17.21}_{\pm2.68}$ (\textcolor{blue}{4.90}) & \textcolor{blue}{2.72} \\
    \rowcolor{Gray}
    \textbf{\oursref-UPGrad} & ${6.47}_{\pm2.46}$ (\textcolor{blue}{2.34}) & ${99.38}_{\pm0.21}$ (\textcolor{blue}{0.62}) & ${92.38}_{\pm0.63}$ (\textcolor{blue}{1.96}) & ${15.31}_{\pm3.52}$ (\textcolor{blue}{4.00}) & \textcolor{blue}{2.23} \\
    \midrule
    \multicolumn{6}{c}{\textbf{Random 50\%}} \\
    \midrule
    \textbf{Methods} & \multicolumn{1}{c|}{UA $\uparrow$} & \multicolumn{1}{c|}{RA $\uparrow$} & \multicolumn{1}{c|}{TA $\uparrow$} & \multicolumn{1}{c|}{MIA $\uparrow$} & \multicolumn{1}{c}{Avg. Gap $\downarrow$} \\
    \midrule
    \rowcolor{white}
    {{\retrain}} & ${19.54}_{\pm0.00}$ (\textcolor{blue}{0.00}) & ${100.00}_{\pm0.00}$ (\textcolor{blue}{0.00}) & ${79.74}_{\pm0.00}$ (\textcolor{blue}{0.00}) & ${20.09}_{\pm0.00}$ (\textcolor{blue}{0.00}) & \textcolor{blue}{0.00} \\
    \midrule
    \FT & ${3.89}_{\pm5.65}$ (\textcolor{blue}{15.65}) & ${97.77}_{\pm3.53}$ (\textcolor{blue}{2.23}) & ${91.71}_{\pm4.21}$ (\textcolor{blue}{11.97}) & ${6.87}_{\pm8.48}$ (\textcolor{blue}{13.22}) & \textcolor{blue}{10.77} \\
    \GA & ${55.82}_{\pm51.09}$ (\textcolor{blue}{51.91}) & ${44.28}_{\pm51.00}$ (\textcolor{blue}{55.72}) & ${42.19}_{\pm47.97}$ (\textcolor{blue}{49.44}) & ${36.42}_{\pm48.98}$ (\textcolor{blue}{39.66}) & \textcolor{blue}{49.19} \\
    \IU & ${20.49}_{\pm26.35}$ (\textcolor{blue}{20.85}) & ${79.75}_{\pm26.11}$ (\textcolor{blue}{20.25}) & ${74.97}_{\pm24.05}$ (\textcolor{blue}{18.05}) & ${19.81}_{\pm22.20}$ (\textcolor{blue}{17.45}) & \textcolor{blue}{19.15} \\
    \MUSparse & ${45.12}_{\pm43.20}$ (\textcolor{blue}{41.40}) & ${54.96}_{\pm43.06}$ (\textcolor{blue}{45.04}) & ${52.32}_{\pm40.17}$ (\textcolor{blue}{38.79}) & ${42.01}_{\pm38.21}$ (\textcolor{blue}{32.36}) & \textcolor{blue}{39.40} \\
    \RL & ${28.29}_{\pm36.72}$ (\textcolor{blue}{30.28}) & ${72.15}_{\pm37.09}$ (\textcolor{blue}{27.85}) & ${68.53}_{\pm33.88}$ (\textcolor{blue}{27.28}) & ${15.00}_{\pm16.35}$ (\textcolor{blue}{14.48}) & \textcolor{blue}{24.97} \\
    SalUn & ${10.36}_{\pm22.06}$ (\textcolor{blue}{21.28}) & ${89.92}_{\pm22.09}$ (\textcolor{blue}{10.08}) & ${84.84}_{\pm19.25}$ (\textcolor{blue}{16.83}) & ${10.99}_{\pm10.82}$ (\textcolor{blue}{13.08}) & \textcolor{blue}{15.32} \\
    \midrule
    \rowcolor{Gray}
    \textbf{\oursuniform-LS} & ${1.59}_{\pm1.73}$ (\textcolor{blue}{17.95}) & ${99.05}_{\pm1.11}$ (\textcolor{blue}{0.95}) & ${92.49}_{\pm1.54}$ (\textcolor{blue}{12.75}) & ${8.93}_{\pm2.81}$ (\textcolor{blue}{11.16}) & \textcolor{blue}{10.70} \\
    \rowcolor{Gray}
    \textbf{\oursuniform-UPGrad} & ${11.23}_{\pm6.69}$ (\textcolor{blue}{8.31}) & ${90.63}_{\pm5.51}$ (\textcolor{blue}{9.37}) & ${84.78}_{\pm5.23}$ (\textcolor{blue}{5.04}) & ${17.23}_{\pm6.35}$ (\textcolor{blue}{4.28}) & \textcolor{blue}{6.75} \\
    \rowcolor{Gray}
    \textbf{\oursref-LS} & ${3.92}_{\pm2.06}$ (\textcolor{blue}{15.61}) & ${98.21}_{\pm0.79}$ (\textcolor{blue}{1.79}) & ${90.72}_{\pm1.19}$ (\textcolor{blue}{10.98}) & ${10.37}_{\pm4.52}$ (\textcolor{blue}{9.72}) & \textcolor{blue}{9.52} \\
    \rowcolor{Gray}
    \textbf{\oursref-UPGrad} & ${1.97}_{\pm2.19}$ (\textcolor{blue}{17.57}) & ${98.89}_{\pm1.29}$ (\textcolor{blue}{1.11}) & ${92.12}_{\pm1.95}$ (\textcolor{blue}{12.38}) & ${14.41}_{\pm6.04}$ (\textcolor{blue}{5.83}) & \textcolor{blue}{9.22} \\
    \midrule
    \bottomrule[1pt]
    \end{tabular}%
    }
    \vspace*{-3mm}
\end{table}

\subsection{Results and Analysis}
Table~\ref{tab: classification_data_ratio} reports the performance of all methods under 10\% and 50\% random forgetting on CIFAR-10 with ResNet-18, measured by Unlearning Accuracy (UA), Retain Accuracy (RA), Test Accuracy (TA), and Membership Inference Attack accuracy (MIA). Our four {\ours} variants achieve the smallest average gaps to {\retrain} in both settings and outperform every baseline we consider.

At 10\% forgetting, {\oursref} with UPGrad aggregator reaches the best average gap of 2.23, followed by {\oursuniform} with UPGrad at 2.53, {\oursref} with LS at 2.72, and {\oursuniform} with LS at 3.24. The strongest baselines are {\IU} at 6.06 and {\FT} at 8.44, while {\GA} reaches 16.90 and {\MUSparse} collapses to 44.15. These results indicate that error-maximization baselines tend to damage retention or fail to forget reliably. Aligning forget predictions with a reference distribution produces a better balance.

The privacy metrics further support our approach. The MIA scores of {\oursref} with UPGrad and {\oursref} with LS are 15.31 and 17.21, which are close to the {\retrain} value of 13.13. In contrast, {\GA} drops to 1.93 and {\MUSparse} inflates MIA to 21.43. Keeping MIA close to the retraining target is a strong signal that {\ours} removes sample influence without the instability of adversarial forgetting.

The 50\% forgetting scenario exposes the fragility of the baselines. {\GA} collapses with an RA of 44.28 and a TA of 42.19, producing a gap of 49.19. {\MUSparse} and {\RL} also deteriorate with gaps of 39.40 and 24.97. {\oursuniform} with UPGrad achieves the best gap of 6.75 and outperforms every baseline by a wide margin. {\oursref} with UPGrad and {\oursref} with LS follow with 9.22 and 9.52, while {\FT} sits at 10.77.

The two {\ours} families reveal a complementary trade-off. {\oursref} with UPGrad preserves high retention with an RA of 98.89 and a TA of 92.12 while using a conservative UA of 1.97, which makes it suitable when protecting retain accuracy matters most. {\oursuniform} with UPGrad offers a more aggressive forgetting profile with UA 11.23 and RA 90.63, and it attains the lowest average gap. The standard deviations of all {\ours} variants are also markedly lower than those of {\GA}, {\MUSparse}, and {\RL}. This shows that reference-aligned forgetting produces stable updates even when half of the training data must be removed.

\paragraph{Effect of the gradient aggregator.} A direct comparison of UPGrad against LS within each {\ours} configuration isolates the effect of the aggregator. At 10\% forgetting, UPGrad reduces the average gap from 3.24 to 2.53 for {\oursuniform} and from 2.72 to 2.23 for {\oursref}. At 50\% forgetting, UPGrad reduces the gap from 10.70 to 6.75 for {\oursuniform} and from 9.52 to 9.22 for {\oursref}. The improvement is consistent across both reference strategies and both forgetting ratios. The numerical gains together with the lower variance of UPGrad runs suggest that the Pareto-stationary solutions reached by UPGrad correspond to better approximate unlearning behavior with smaller gaps to {\retrain}.

\section{Conclusion}
We presented {\ours}, a framework that formulates machine unlearning as a bi-objective optimization problem and replaces adversarial loss maximization with alignment to a reference distribution representing unseen data. This reformulation reduces the gradient conflict between forgetting and retention and enables stable updates under Jacobian descent with the UPGrad aggregator. The two reference strategies offer a complementary advantage. {\oursuniform} needs no extra data and delivers stronger forgetting, while {\oursref} delivers tighter retention when a held-out set is available.
Since {\oursref} may not be available in all practical scenarios, {\oursuniform} serves as a fallback. Constructing reference distributions from limited or synthetic data is an important direction for broader applicability. 

In future work, we plan to extend multi-objective {\ours} to generative models and LLMs. We also plan to study MOEAs \cite{deb2002fast} as alternatives to Jacobian descent, which produce more diverse Pareto-optimal solutions for unlearning.

\section*{Acknowledgement}
This work is supported by Natural Sciences and Engineering Research Council of Canada.
\printbibliography[heading=subbibintoc]

@misc{gdpr2016,
  title = {{Regulation (EU) 2016/679 of the  European Parliament and of the Council}},
  author       = {{European Parliament and Council of the European Union}},
  year         = {2016},
  howpublished = {\textit{Official Journal of the European Union, L119}, 1--88}
}

@misc{CatastrophicForgetting,
title={Overcoming catastrophic forgetting beyond continual learning: Balanced training for neural machine translation},
author={Shao, Chenze and Feng, Yang},
journal={arXiv preprint arXiv:2203.03910},
year={2022}
}

@inproceedings{UnlearnPretrainedLLM,
    title = "Machine Unlearning of Pre-trained Large Language Models",
    author = "Yao, Jin  and
      Chien, Eli  and
      Du, Minxin  and
      Niu, Xinyao  and
      Wang, Tianhao  and
      Cheng, Zezhou  and
      Yue, Xiang",
    editor = "Ku, Lun-Wei  and
      Martins, Andre  and
      Srikumar, Vivek",
    booktitle = "Proceedings of the 62nd Annual Meeting of the Association for Computational Linguistics (Volume 1: Long Papers)",
    month = aug,
    year = "2024",
    address = "Bangkok, Thailand",
    publisher = "Association for Computational Linguistics",
    doi = "10.18653/v1/2024.acl-long.457",
    pages = "8403--8419"
}

@inproceedings{MUNBAUnlearning,
  title={Munba: Machine unlearning via nash bargaining},
  author={Wu, Jing and Harandi, Mehrtash},
  booktitle={Proceedings of the IEEE/CVF International Conference on Computer Vision},
  pages={4754--4765},
  year={2025}
}

@misc{GSLoRApracticalforgetting,
title={Continual forgetting for pre-trained vision models},
author={Zhao, Hongbo and Ni, Bolin and Fan, Junsong and Wang, Yuxi and Chen, Yuntao and Meng, Gaofeng and Zhang, Zhaoxiang},
booktitle={Proceedings of the IEEE/CVF Conference on Computer Vision and Pattern Recognition},
pages={28631--28642},
year={2024}
}

@misc{MOO-LLMUnlearning,
title={Multi-Objective Large Language Model Unlearning},
author={Pan, Zibin and Zhang, Shuwen and Zheng, Yuesheng and Li, Chi and Cheng, Yuheng and Zhao, Junhua},
booktitle={ICASSP 2025-2025 IEEE International Conference on Acoustics, Speech and Signal Processing (ICASSP)},
pages={1--5},
year={2025},
organization={IEEE}
}

@misc{cheng2024remainingdatafreemachineunlearningsuppressing,
title={Remaining-data-free machine unlearning by suppressing sample contribution},
author={Cheng, Xinwen and Huang, Zhehao and Zhou, Wenxin and He, Zhengbao and Yang, Ruikai and Wu, Yingwen and Huang, Xiaolin},
journal={arXiv preprint arXiv:2402.15109},
year={2024}
}

@article{JacobianDescent,
  title={{Jacobian Descent For Multi-Objective Optimization}},
  author={Quinton, Pierre and Rey, Valérian},
  journal={arXiv preprint arXiv:2406.16232},
  year={2024}
}

@article{SalUn,
  title={Salun: Empowering machine unlearning via gradient-based weight saliency in both image classification and generation},
  author={Fan, Chongyu and Liu, Jiancheng and Zhang, Yihua and Wong, Eric and Wei, Dennis and Liu, Sijia},
  journal={arXiv preprint arXiv:2310.12508},
  year={2023}
}

@article{zhao2024makes,
  title={{What makes unlearning hard and what to do about it}},
  author={Zhao, Kairan and Kurmanji, Meghdad and B{\u{a}}rbulescu, George-Octavian and Triantafillou, Eleni and Triantafillou, Peter},
  journal={Advances in Neural Information Processing Systems},
  volume={37},
  pages={12293--12333},
  year={2024}
}

@inproceedings{thudi2022unrolling,
  title={Unrolling sgd: Understanding factors influencing machine unlearning},
  author={Thudi, Anvith and Deza, Gabriel and Chandrasekaran, Varun and Papernot, Nicolas},
  booktitle={2022 IEEE 7th European Symposium on Security and Privacy (EuroS\&P)},
  pages={303--319},
  year={2022},
  organization={IEEE}
}

@inproceedings{izzo2021approximate,
  title={Approximate data deletion from machine learning models},
  author={Izzo, Zachary and Smart, Mary Anne and Chaudhuri, Kamalika and Zou, James},
  booktitle={International Conference on Artificial Intelligence and Statistics},
  pages={2008--2016},
  year={2021},
  organization={PMLR}
}

@inproceedings{golatkar2020eternal,
  title={Eternal sunshine of the spotless net: Selective forgetting in deep networks},
  author={Golatkar, Aditya and Achille, Alessandro and Soatto, Stefano},
  booktitle={Proceedings of the IEEE/CVF Conference on Computer Vision and Pattern Recognition},
  pages={9304--9312},
  year={2020}
}

@article{warnecke2021machine,
  title={Machine Unlearning of Features and Labels},
  author={Warnecke, Alexander and Pirch, Lukas and Wressnegger, Christian and Rieck, Konrad},
  journal={arXiv preprint arXiv:2108.11577},
  year={2021}
}

@article{Li2023RandomRelabeling,
  title={Random Relabeling for Efficient Machine Unlearning},
  author={Li, Junde and Ghosh, Swaroop},
  journal={arXiv preprint arXiv:2305.12320},
  year={2023}
}

@article{jia2023model,
  title={Model sparsification can simplify machine unlearning},
  author={Jia, Jinghan and Liu, Jiancheng and Ram, Parikshit and Yao, Yuguang and Liu, Gaowen and Liu, Yang and Sharma, Pranay and Liu, Sijia},
  journal={arXiv preprint arXiv:2304.04934},
  year={2023}
}

@article{desideri2012mgda,
  title={Multiple-gradient descent algorithm (MGDA) for multiobjective optimization},
  author={D{\'e}sid{\'e}ri, Jean-Antoine},
  journal={Comptes Rendus Mathematique},
  volume={350},
  number={5-6},
  pages={313--318},
  year={2012},
  publisher={Elsevier}
}

@inproceedings{sener2018multitask,
  title={Multi-task learning as multi-objective optimization},
  author={Sener, Ozan and Koltun, Vladlen},
  booktitle={Advances in Neural Information Processing Systems},
  volume={31},
  year={2018}
}

@article{deb2002fast,
  title={A fast and elitist multiobjective genetic algorithm: NSGA-II},
  author={Deb, Kalyanmoy and Pratap, Amrit and Agarwal, Sameer and Meyarivan, TAMT},
  journal={IEEE transactions on evolutionary computation},
  volume={6},
  number={2},
  pages={182--197},
  year={2002},
  publisher={Ieee}
}

\newpage

\appendix

\section{Experimental Setup}
\label{app:setup}
We evaluate {\ours} on CIFAR-10 using ResNet-18 under two random-data forgetting scenarios with 10\% and 50\% of the training set removed (4{,}500 and 22{,}500 samples, respectively), starting from the same pretrained checkpoint in all runs. The reference {\retrain} model is trained from scratch for 182 epochs with an initial learning rate of $10^{-1}$ and a multi-step learning rate schedule that multiplies the learning rate by a factor $\gamma = 0.1$ at epochs 91 and 136. We compare against six unlearning baseline methods: Fine-Tuning ({\FT})~\cite{warnecke2021machine}, Gradient Ascent ({\GA})~\cite{thudi2022unrolling}, Influence Unlearning ({\IU})~\cite{izzo2021approximate,jia2023model}, {\MUSparse}~\cite{jia2023model}, Random Labeling ({\RL})~\cite{golatkar2020eternal}, and SalUn~\cite{SalUn} (random-label unlearning with a fixed saliency mask). All methods use SGD as the optimizer with fixed learning rate (no learning rate scheduler). Except for {\GA}, which uses 2 epochs, baselines use 10 unlearning epochs. Learning rates are fixed per run according to the grids as follows. For {\FT}, the five runs use learning rates $\{10^{-1},\,10^{-2},\,5\times 10^{-2},\,5\times 10^{-3},\,5\times 10^{-3}\}$. For {\RL}, they are $\{10^{-1},\,5\times 10^{-2},\,10^{-2},\,5\times 10^{-3},\,10^{-3}\}$. For {\GA}, LRs are $\{10^{-4},\,5\times 10^{-4},\,10^{-3},\,5\times 10^{-3},\,7\times 10^{-3},\,9\times 10^{-3},\,10^{-2},\,2\times 10^{-2}\}$ at 10\% forgetting and $\{5\times 10^{-4},\,10^{-3},\,5\times 10^{-3},\,10^{-2},\,2\times 10^{-2}\}$ at 50\% forgetting. {\IU} uses learning rate $10^{-3}$ with influence coefficient $\alpha \in \{1, 5, 10, 15, 20\}$. {\MUSparse} sweeps $\{10^{-6},\,5\times 10^{-6},\,10^{-5},\,2\times 10^{-5},\,3\times 10^{-5},\,5\times 10^{-5},\,10^{-4}\}$. SalUn sweeps $\{10^{-4},\,5\times 10^{-4},\,10^{-3},\,10^{-2},\,5\times 10^{-2}\}$. 

We configure our multi-objective {\ours} framework with two gradient aggregators: fixed linear scalarization (LS) and UPGrad~\cite{JacobianDescent}. We use SGD as the optimizer with learning rates sweeping in $\{10^{-3},\,2.5\times 10^{-3},\,5\times 10^{-3},\,7.5\times 10^{-3},\,10^{-2},\,5\times 10^{-2}\}$ without any learning rate scheduling. 

We report UA, RA, TA, and MIA following the benchmarking protocol of~\cite{SalUn}. Performance is measured by the gap to {\retrain}, and a smaller gap indicates better approximate unlearning.

\end{document}